\documentclass[shortpaper]{clv2}
\usepackage{times}
\usepackage{url}
\usepackage{latexsym}
\usepackage{graphicx}
\usepackage{amsmath}
\usepackage{subcaption}
\usepackage{color}

\title{Improving Automated Patent Claim Parsing: Dataset, System, and Experiments}

\author{Mengke Hu, David Cinciruk, and John MacLaren Walsh}
\affil{Dept. of Elec. \& Comp. Eng., Drexel University, Philadelphia, PA}

\begin{document}
\maketitle
\begin{abstract}
Off-the-shelf natural language processing software performs poorly when parsing patent claims owing to their use of irregular language relative to the corpora built from news articles and the web typically utilized to train this software.
Stopping short of the extensive and expensive process of accumulating a large enough dataset to completely retrain parsers for patent claims, a method of adapting existing natural language processing software towards patent claims via forced part of speech tag correction is proposed.  An Amazon Mechanical Turk collection campaign organized to generate a public corpus to train such an improved claim parsing system is discussed, identifying lessons learned during the campaign that can be of use in future NLP dataset collection campaigns with AMT.  Experiments utilizing this corpus and other patent claim sets measure the parsing performance improvement garnered via the claim parsing system.  Finally, the utility of the improved claim parsing system within other patent processing applications is demonstrated via experiments showing improved automated patent subject classification when the new claim parsing system is utilized to generate the features.
\end{abstract}

\section{Introduction}

Prior research has demonstrated that natural language processing (NLP) based features, such as dependency lists \cite{[176]dependencies}, can provide useful features for the automated processing of patents \cite{[167]textRepPatent}, owing to their ability to better capture deeper semantic relationships than n-grams and simple term and document frequency based features typically used in text classification and information retrieval.  Because the claims of a patent define the invention and its scope from a legal standpoint, they contain a substantial amount of the information in a patent that is germane to these patent processing problems, which include, among others, subject classification, prior art search, valuation, infringement analysis, and patent information retrieval \cite{2013ChallengePatentRetrieval,[56]MihaiLupu_StatusRetrievalEval_2011_PAIR,[153]fntir_2013}.  However, patent claims are especially difficult for off-the-shelf NLP parsers such as the Stanford Parser \cite{StanfordParser} to parse \cite{[161]QuantifyingTheChallengesInParsingPatentClaims,[78]KristineAtkinson_RationalPatentSearch_2008_PAIR}.  These NLP engines utilize language models trained from news article or web corpora such as the Wall Street Journal corpus \cite{StanfordDependencies}, which are poorly matched to the otherwise atypical language utilized in patent claims.

This mismatch arises from a number of factors, some of which have been documented \cite{[161]QuantifyingTheChallengesInParsingPatentClaims,[78]KristineAtkinson_RationalPatentSearch_2008_PAIR}.  To begin with, patent claims involve sentences that are much longer than those typically found in other sources of text.  This is because in each claim, patent agents and attorneys must be able to cover all possible forms of the invention in one run-on sentence with very precise language.  A simple method adopted by the patent parsing community to partially addresses this problem is to chunk the long patent claims into smaller segments \cite{[69]PeterParapatics_PatentClaimDecompo_2009_PAIR}.  This method also helps to avoid the time and memory requirements to parse long sentences like most patent claims.

\begin{figure*}[t!]
    \centering
    \begin{subfigure}[t]{1\textwidth}
        \centering
        \includegraphics[width=0.9\textwidth]{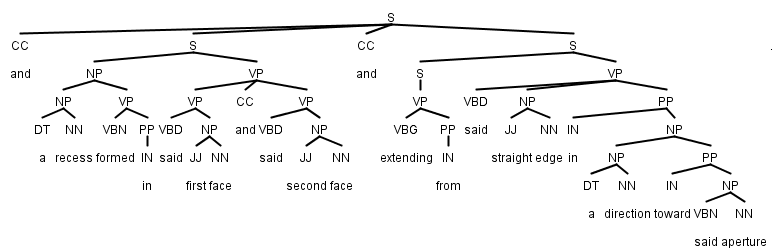}
        \caption{Incorrect Parse Tree of a Patent Claim Segment}
        \label{fig:WrongParse}
    \end{subfigure}
    \begin{subfigure}[t]{1\textwidth}
        \centering
        \includegraphics[width=0.9\textwidth]{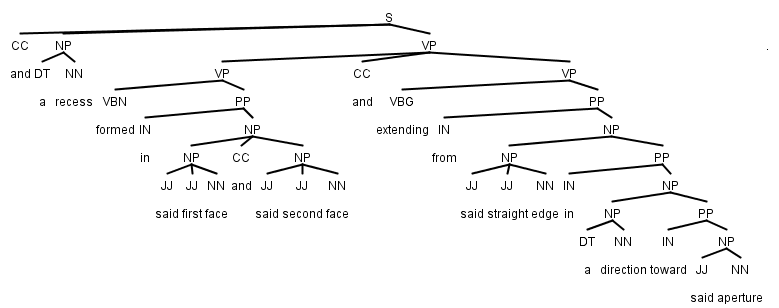}
        \caption{Correct Parse Tree of a Patent Claim Segment}
        \label{fig:CorrectParse}
    \end{subfigure}
    \caption{Originally Incorrect and Fixed Parse Trees for a Patent Claim Segment}
\end{figure*}

Another source of mismatch, which remains after chunking, arises from the use of words as less common parts of speech in patent claim language.  For instance ``said'' most commonly functions as an adjective, and ``claim'' typically functions as a noun, when occurring in patent claims, while these words more typically appear as verbs in news articles, the web, and other sources of text.  Verbs play pivotal roles in parse trees, hence misidentifying other words as acting as verbs substantially damages the ability to construct the correct tree.  This leads to incorrect claim parse trees such as the one produced by the Stanford parser in Fig. \ref{fig:WrongParse}.

The cleanest, first-principles oriented, method to guarantee a better claim parsing is to train a special grammatical model for patents or patent claims and use that model in the parser \cite{[161]QuantifyingTheChallengesInParsingPatentClaims}.  However, in order to train this model, a large hand-annotated corpus of patents or patent claims must be created and utilized.  Creating this corpus requires extensive development time and resources and is thus infeasible and impractical if another solution can be found.

A simpler solution, motivated by the mismatch problem depicted in Fig. \ref{fig:WrongParse}, observes that some NLP packages, such as the Stanford Parser, provide a tool to force certain words to be tagged with certain parts of speech (POS) tags.  As demonstrated in Fig. \ref{fig:CorrectParse}, by just correcting the incorrect verbs tags and rerunning the parser, a correct parsing can often be obtained.  This narrows the attention from the very difficult problem of gathering enough data to retrain an entire grammar to collecting enough data to enable the verification and correction of POS tags for words originally tagged as verbs.  The remainder of this article describes the creation and evaluation of such a POS tag correction based improved claim parser. 

The first order of business was to assemble a dataset from which to train and evaluate the system.  Following an increasing trend in speech and language processing, we crowd-sourced the formation and labeling of this corpus via Amazon Mechanical Turk.  Sec. \ref{sec:AMT} reviews the campaign under which several generations of data were collected, indicating lessons learned about Turker incentives and capabilities that may be of use in future NLP data campaigns of a similar ilk.  Also, as detailed in Sec. \ref{sec:AMT}, the dataset we collected has also been made publicly available for other researchers to train alternative systems.

With the newly formed corpus in hand, Sec. \ref{sec:system} describes the development and testing of a machine learning based system for correcting POS tags for patent claim terms initially tagged as verbs.
Next, the improvement in claim parsing enabled by forcing the corrected POS tags, both on testing data gathered from the AMT campaign, and other patent data, is measured in experiments described in Sec. \ref{sec:eval}.  The improvement in claim parsing enabled by the developed system is further demonstrated in Sec. \ref{sec:subjClass}, where it is shown utilizing the improved parsing system to generate dependency lists for features for patent subject classification substantially improves classification performance over a baseline system using unmodified off-the-shelf NLP tools.

\section{Dataset Collection via Amazon Mechanical Turk}
\label{sec:AMT}

\begin{table*}[t!]
\footnotesize
\centering
\begin{tabular}{| l | l | l | l | l | l | l | l | l | l |}
\hline
Campaigns & \# of Patents & \# of Segments & \# of tag queries & Accept:Reject & Avg. Time/HIT & Pay \\
\hline
Campaign 1 & 10 & 360 & 315 & 58:28 & 3 min 45 sec  & \$0.05 \\
Campaign 2 & 200 & 6690 & 8983 & 311:357 & 12 min 52 sec& \$0.45\\
Campaign 3 & 591 & 24560 & 62666 & 638:589 & 18 min 52 sec & \$1.20 \\
Campaign 4 & 888 & 39021 & 141920 & 1774:1142 & 11min 56 sec & \$0.50 \\
\hline
\end{tabular}
\caption{Patent claim data composition of the four AMT data collection campaigns of broadening scope and scale. }
\label{tab:AMTExperiments}
\end{table*}

Amazon Mechanical Turk is a crowdsourcing marketplace that researchers have used to gather data for their experiments and systems.  Each project posted on AMT by a researcher allows that researcher to obtain data by paying workers a small amount of money to perform a small amount of work via a Human Intelligence Task (HIT). In this section, inspired by papers such as \cite{[249]CreatingSpeechLanguage_AMT,[246]RankingAnnotator_AMT,[247]QualityControl_AMT,[238]crowdsourcingGrammatical,[233]AMTannotation}, we describe how we used AMT to collect correct POS tags from words originally labeled as verbs and what we observed from the results.

As explained in the introduction, previously, during an exploratory test, we had determined that the Stanford Parser would routinely tag certain patent claim terms as verbs even though they are actually different parts of speech.  After correcting these incorrect tags and re-running the Stanford parser forcing these corrected tags, we were able to generate parse trees that more accurately represent the patent claim.  It was from this test that we determined that we could see improvements by just having workers correct the tags of putative verbs, i.e. correct/verify the POS tags of terms that were labelled as verbs after an initial run of the Stanford NLP parser.  This sort of task was also deemed more well within the expertise of the pool of talent available through mechanical turk than, say, providing POS tags for all terms, or, even harder, providing an entire parse tree annotation.

The tag correction dataset collection was carried out in four campaigns, each of increasing scope and efficiency in rate and cost of curated (useful) data collection. An overview of these four campaigns, indicating the composition and growing scope of the data they annotated, as well as raw statistics about their human intelligence task (HIT) units is provided in Table \ref{tab:AMTExperiments}.  As outlined in this table, for each campaign, the claims from a number of randomly selected patents were broken into segments at punctuation, and processed with the Stanford parser to obtain a series of initial part of speech (POS) tags.  Turkers were asked to verify if those words which were tagged as verbs were tagged correctly, and, if not, to tag the word as acting as a noun, adjective, adverb, or ``other''.  

As many authors have documented, crowdsourcing dataset generation requires careful curation, since the workers have a variety of educational backgrounds and have direct financial incentives to rapidly perform the task by not performing it well if they can get paid for it.  We addressed this issue by including in each of the campaigns test questions that were chosen randomly from a pool of about 100 different test patent claim segments.  After receiving responses from turkers, we ran them through a program that compared the answers of the test questions with a list of acceptable answers determined by hand.  Turkers would have their HIT accepted only if they provided sufficiently accurate answers on the test questions.   Only these accepted HITs were included in the subsequent curation process to create the tag dataset.  As can be seen in Table \ref{tab:AMTExperiments}, uniformly across the tasks, a substantial fraction of responses were not able to meet the threshold for accuracy of responses on these test questions, indicating that the task was difficult, but became easier as the campaigns evolved.

The first three campaigns were the easiest to generate HITs for, and collected data already in a format easily utilized with natural language processing software, but utilized interfaces that were substantially less user friendly than the fourth.  The only differences among these first three campaigns were the number of segments in each HIT and the associated advertised pay per HIT.  Conversely, the fourth campaign's HIT interface enabled workers to more easily and rapidly complete the task, but required a more complicated backend task organization and data curation process.  

\subsection{Three Initial Tag Collection Campaigns}

\begin{figure*}[t!]
    \centering
    \begin{subfigure}[t]{1\textwidth}
        \centering
        \includegraphics[width=0.9\textwidth]{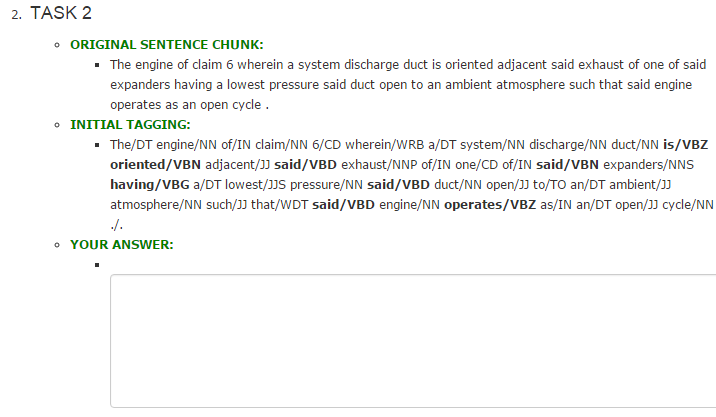}
        \caption{Sample Question from our First Three AMT Campaigns}
        \label{fig:AMT123Question}
    \end{subfigure}
    \begin{subfigure}[t]{1\textwidth}
        \centering
        \includegraphics[width=0.9\textwidth]{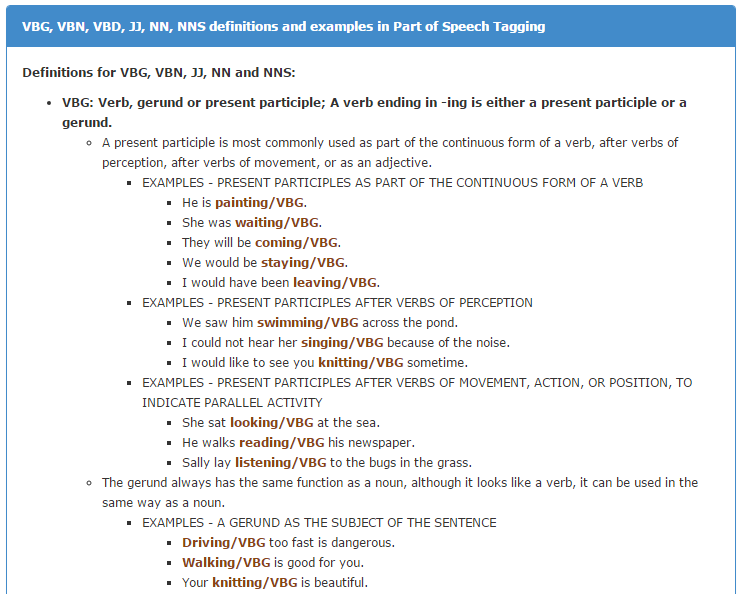}
        \caption{Portion of the Instructions from our First Three AMT Campaigns}
        \label{fig:AMT123Instructions}
    \end{subfigure}
    \caption{Question and Instructions from our First Three AMT Campaigns}
\end{figure*}

In the first three campaigns, whose query interface is depicted in Fig. \ref{fig:AMT123Question}, the workers were given an original sentence segment followed by the initial POS tagged version of it, then were tasked with changing any incorrect \textbackslash VB* tags to either \textbackslash NN, or \textbackslash JJ.  While this task was designed to be conceptually easy to grasp, the amount of unnecessary tags included in the sentence and the sheer length of some of these sentences made the task difficult.  As reading fatigue accumulated, it was not uncommon for workers to overlook verbs requiring correction at the end of sentences.

This fatigue was compounded by the necessity of a large set of instructions required to enable the turkers to understand the POS tag encoding and the format in which their responses were to be encoded.  A screenshot of a part of the instructions could be seen in Fig. \ref{fig:AMT123Instructions}.  Due to the relative unfamiliarity most workers had with the output of the Stanford POS Tagger, we had to give every worker an overview of the POS tagged sentences.  We also provided definitions (and examples) of all the tags we were interested in.  In addition to that, the workers were provided with already-completed examples to show them what we wanted as an input.

For our initial campaign, the results trickled in at such a slow pace that we could have tagged the data far faster by hand ourselves: as can be seen from Table \ref{tab:efficiency}, only three new part of speech tags were being generated for every working hour.  A natural initial suspicion was that the low HIT payment had dissuaded workers from reading and accepting our HITs.  So, as shown in Table \ref{tab:AMTExperiments} we raised the number of segments we had workers correct from 2 (1 real and 1 test) to 10 (9 real and 1 test) and raised the payment for successful completion of a HIT from 5 cents to 45 cents.  In this manner, the payment per new curated data segment roughly doubled, but the reward per hit increased by nearly a factor of 10.  This increased the level of interest, yielding many more turkers responding to the task, and increased the rate at which data was being tagged to 42 tags per workday hour (c.f. Table \ref{tab:efficiency}).  While this represented an tenfold annotation rate improvement, it still was slower than performing the tagging ourselves.  However, this marked increase in data collection rate suggested that the turkers were especially sensitive to the total reward per hit, perhaps because when users arrive at the site it is likely that many of them rank the HITs by total reward per HIT.  Bearing this in mind, in the third campaign we increased the number of segments in each HIT to 20 (18 real and 2 test) and raised the payment per HIT to \$1.20, selected to place us on the first few pages of the HIT list when ranked by pay per HIT.  As a result, the rate at which we were receiving newly annotated tags increased by a factor of five to a more respectable 243 tags per workday hour, even though we were paying slightly less per curated tag tag than we had in the previous campaign.

Also of interest was that over the evolution of the first three campaigns the average rate that individual turkers completed curated tags steadily increased, even though the user interface remained somewhat static, with only a variation in the number of queries (see the average accepted worker tag rate in Table \ref{tab:efficiency}).  Several factors are likely to have to increase in worker effectiveness.   The first is that as the number responses included in each HIT grew, the initial time loss of reading and understanding the instructions was amortized over a larger and larger number of responses.  A second factor is that a pool of workers contributed to several subsequent campaigns, and thus became accustomed to working quickly with the interface.  A third, final, and likely most important factor, is that as the total HIT reward increased, the pool of workers considering and participating in the HIT grew, enabling a wider collection of workers capable of rapidly performing the part of speech tagging accurately to be reached.

\subsection{A Fourth Campaign with an Improved User Interface}

\begin{figure*}[t!]
    \centering
    \begin{subfigure}[t]{1\textwidth}
        \centering
        \includegraphics[width=0.6\textwidth]{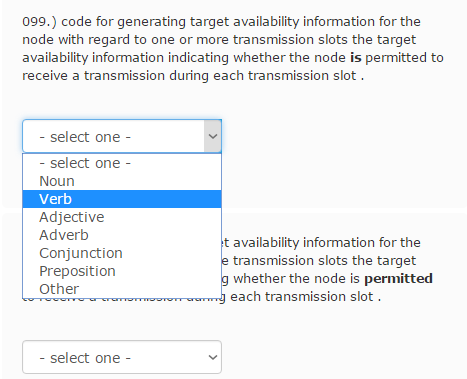}
        \caption{Sample Question from our Fourth AMT Campaigns}
        \label{fig:AMT4Question}
    \end{subfigure}
    \begin{subfigure}[t]{1\textwidth}
        \centering
        \includegraphics[width=0.9\textwidth]{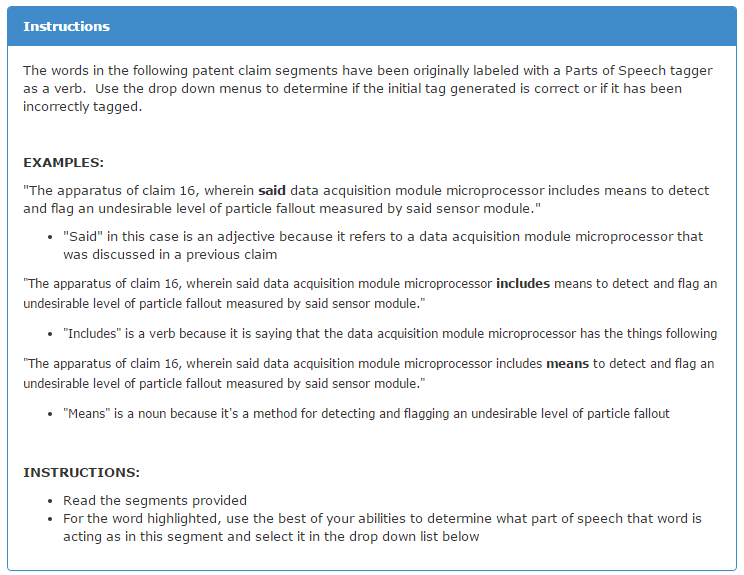}
        \caption{Instructions from our Fourth AMT Campaigns}
        \label{fig:AMT4Instructions}
    \end{subfigure}
    \caption{Question and Instructions from our Fourth AMT Campaigns}
\end{figure*}

However, based on email dialogue and conversations with turkers participating in the first three tasks, it was clear that even though we had increased the response rate to a remotely respectable level, the idiosyncracies of the cut and paste user interface and the unwieldy collection of tags that were not to be corrected rendered the first three campaigns unnecessarily inefficient.  Additionally, something needed to be done to aid the sizable fraction of turkers which were earnestly attempting the task only to have their responses rejected due to accidental errors they made from fatigue.
Hence, for the fourth campaign, we embarked on a redesign of the task to minimize the set of instructions and to maximize the ease of response from the turkers.

The user interface for the fourth campaign that resulted is depicted in Fig. \ref{fig:AMT4Question}.  Instead of giving workers a POS-tagged sentence and having them work with the complete sentence for one task of the HIT, the workers were instead given a sentence (without POS tags) and are asked to only consider the parts of speech of a single bolded verb.    Additionally, to minimize the possibility of typos, workers are given a drop-down list featuring different potential parts of speech that the word could be and are asked to choose the correct parts of speech.  A HIT in this experiment does not contain a set number of segments to correct.  Instead, each HIT consists of exactly 100 verbs, grouped by originating verbs, to correct, of which 20 of them are test questions.  This lead to greater uniformity in the real effort necessary to complete each HIT.

The simpler user interface also lead to simpler instructions, as depicted in Fig. \ref{fig:AMT4Instructions}.  Now the instructions could focus on the essence of the data collection task rather than the mechanics of the response.  In particular, the instructions for the fourth campaign mainly consisted of example segments from patent sentences in which we analyzed each word in a sentence which had been initially tagged as a verb, although this initial tagging was no longer revealed to participants, and explained what the proper tagging among the provided choices should be and why.

As the user interface for this task enabled turkers to more rapidly respond, it also opened the possibility for rapid random responses, hence, testing for correctness became more involved in this campaign than the previous three.  20 test questions were randomly interspersed among the 100 queries.  Initially, as with the previous experiments, we required correct answers to all of these twenty questions in order to accept the answers, but after hearing grievances from those that got most, but not all, correct, we reduced the number of required correct answers from 20 to 18, keeping the data and allowing turkers to still get paid if they got 90\% of the test POS tags correct.

Of all of the campaigns, this fourth and final campaign was by far the most successful.  In just 6 days, we gathered over one hundred thousand new POS tag annotations for putative verbs in patent claims spending less than \$900 in total to do so.  Furthermore, the user interface enabled a large increase in the fraction of responses which could be accepted as well as a near doubling in the rate at which individual users could provided useful tags.  It would have required nine traditional employees, not to mention extra computer hardware and other overhead, to annotate data at the rate it was being collected in this final campaign.  Furthermore, curated (i.e. POS tags that were not only part of accepted HITs, but also passed subsequent validation processes) data from this final campaign was collected at the cheapest price, costing less than three quarters of a cent for each POS tag. 

\subsection{Lessons Learned about Collecting POS Tags with AMT}
Several valuable conclusions regarding the collection of natural language processing data with Amazon Mechanical Turk can be gleaned from the experience provided by campaigns 1--4 summarized in key statistics provided in Table \ref{tab:efficiency}.  

When considering the design of HITs for dataset curation, it is important to first consider the goals of the participants.  The requester creating the HITs naturally wants a large number of high quality annotations to be gathered as rapidly as possible and with a low cost.  The turkers, in turn, are motivated by a combination of factors, which include, perhaps most importantly, to be paid as much as possible for their effort, but also interest in the task that the HIT is contributing to.  The most successful campaigns achieve these aims simultaneously for both parties.

Naturally, one expects that if a requester pays more per unit effort, they will interest more workers and get better quality responses.  However, a first lesson learned form our four campaigns is that it is important, if possible, to provide a reward per HIT that enables one to be on one of the first few pages of HITs open to all turkers when ranked by rewards in order to get a robust collection of turkers attempting the HIT.  As the data in Table \ref{tab:efficiency} shows, this need not imply that one must pay more per annotation when doing so: indeed, by putting more work into the individual HITs one can even obtain useful tags at a lower cost while simultaneously getting more users to respond effectively to the HIT.  

A second lesson, even more important than the first, which can be learned by comparing the vast difference between campaign 4 and the first three in Table \ref{tab:efficiency}, is that the most significant factor in determining the success and rate at which data can be annotated is the efficiency and user friendliness of the user interface presented to the turkers.  Instructions should be sufficient, but minimal, and the user interface should require minimal effort from the turkers.  Additionally, when the user interface enables more rapid response, (compare the average accepted worker tag rate with the number of equivalent parallel local employees) many more turkers will respond to the task, yielding a doubly improved response rate.

A third, and final lesson, that can be learned from the four campaigns is that there is a rich pool of expertise available on mechanical turk and willing to participate at any time on natural language processing related annotation tasks.  Interestingly, when one breaks the average worker HIT completion time in Table \ref{tab:AMTExperiments} into the average for HITs that were accepted because the turker answered the test questions accurately and the average for HITs that were rejected due to erroneous answers on the test questions, one finds that these averages are fairly comparable.  This indicates that the majority of the turkers whose responses were rejected were not attempting to slip one past the requester, but rather were making genuine mistakes or answering questions with ambiguous answers.  The additional leeway in making these mistakes provided they were not too numerous, and setting up of the user interface in campaign four to make these mistakes more difficult to make, lead to a far more symbiotic relationship between the requester and turkers, and achieved a sweet spot in which both parties were maximally pleased.

\begin{table*}[t!]
\centering
\includegraphics[width=\textwidth]{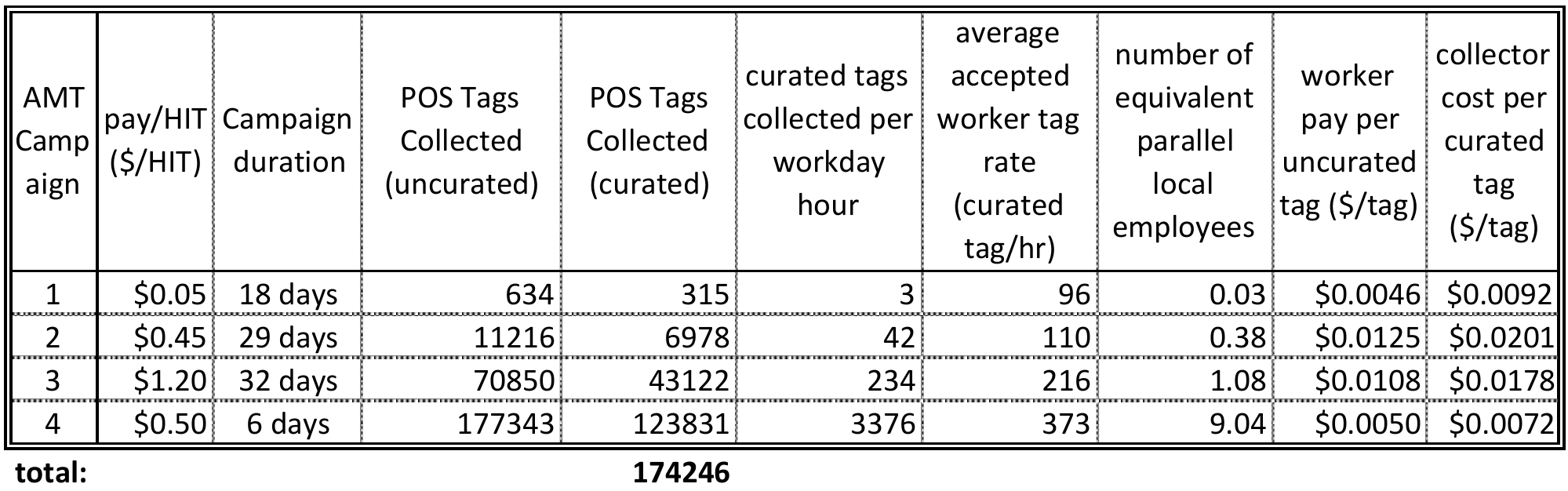}
\caption{Key efficiency statistics for the four AMT POS tag collection campaigns.  Roughly 174,000 new POS tag annotations were collected at a price of roughly a cent per tag.  The campaigns rapidly evolved in efficiency, with the rate of tag collection growing by a factor of 1000, as well as the rate at which individuals provided useful responses growing by a factor of 4.  It would have required nine traditional annotation employees working in parallel to gather data at the rate achieved with the fourth task, which cost a total of just \$882.}\label{tab:efficiency}
\end{table*}

\subsection{Curation of the Dataset}
\label{Curation}

After the accepted HIT responses had been collected, further curation was naturally required to get the data into a form where it could be utilized to train a system for parsing patent claims.  The data from the first three campaigns necessitated the largest amount of curation, as the responses were free-form text, and thus could easily be corrupted with accidental extra spaces or carriage returns in the middle of words, or could mismatch the query entirely.  Additionally, in order to get the text responses into a form usable with standard machine learning paradigms, it was necessary to identify the location of each tagged term (which was previously putatively tagged a verb) within the segment as well as single out the corrected/confirmed tag for it.  After processing the responses with software that reconciled the query with the responses, and manually fixing a few of the responses that were detected as erroneous by the software but could be easily repaired, we obtained the numbers of curated POS tags shown in Table \ref{tab:efficiency}.  This curated data, along with the raw responses from the turk campaigns, have been made available to the public at \cite{AMTPatent}.  Every line in the data set contains a segment of a patent claim, the position within this segment of a term, and the POS tag selected by the turkers for this term.    Overall, across all of the campaigns we collected 174246 unique POS tags.  Additionally, as it is likely that it will be desirable for the training and testing datasets for POS tag correctors to maintain separation between the patents involved, and in curating the data, we have made this easy by including labels for the segments identifying which patents they were generated from.  Analyzing the data, we found that it was important that we had run the campaigns, as only about 69\% of the claim terms originally tagged as a verb by the Stanford parser were actually correctly tagged according to the turkers.

\section{Automatic Parts of Speech Tag Fixer}
\label{sec:system}

\begin{figure}
\centering
\includegraphics[width=.3\textwidth]{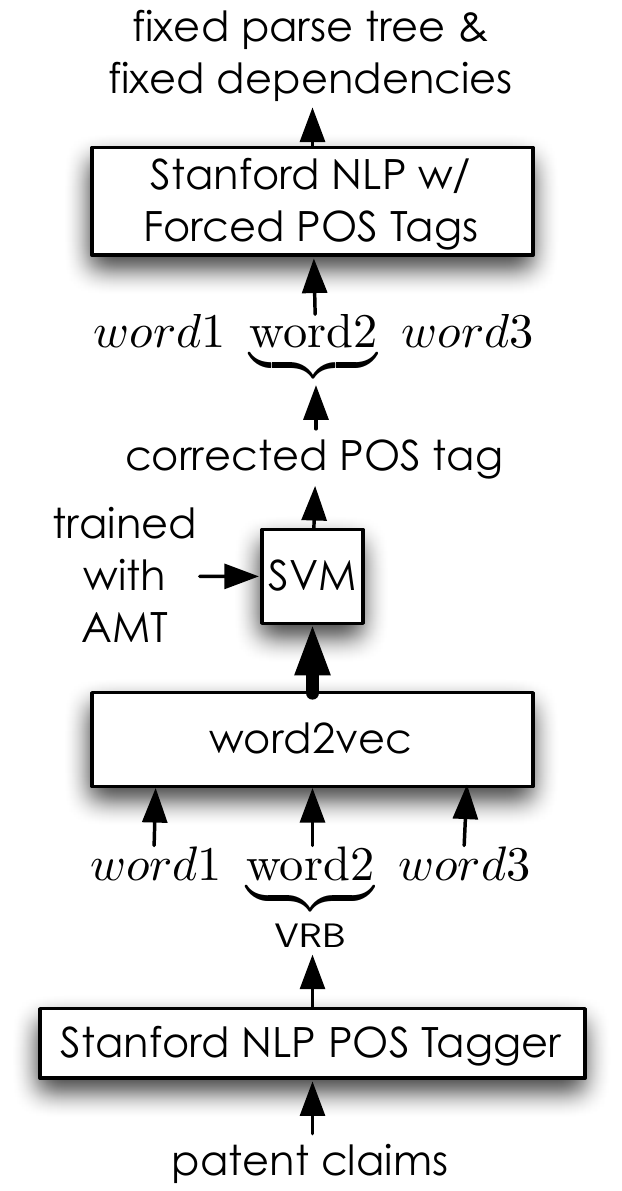}
\caption{Improved claim parsing system via forced POS tag correction.}\label{fig:sysDiag}
\end{figure}

\begin{figure*}[t!]
    \centering
    \begin{subfigure}[t]{0.5\textwidth}
        \centering
        \includegraphics[width=1\textwidth]{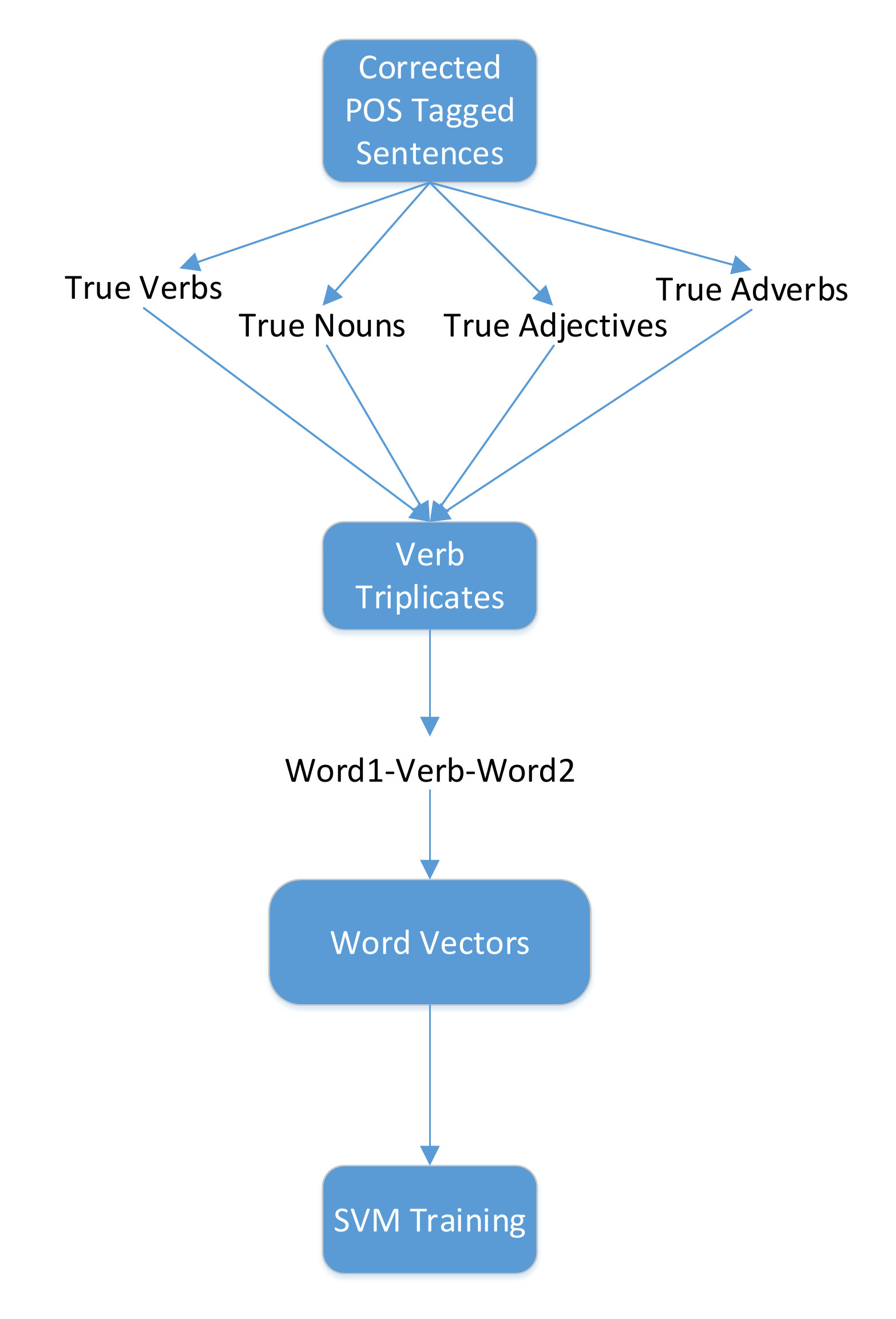}
        \caption{Flowchart of Training Stage of Automatic POS Tag Corrector}
        \label{fig:TrainingStage}
    \end{subfigure}%
    ~
    \begin{subfigure}[t]{0.5\textwidth}
        \centering
        \includegraphics[width=0.9\textwidth]{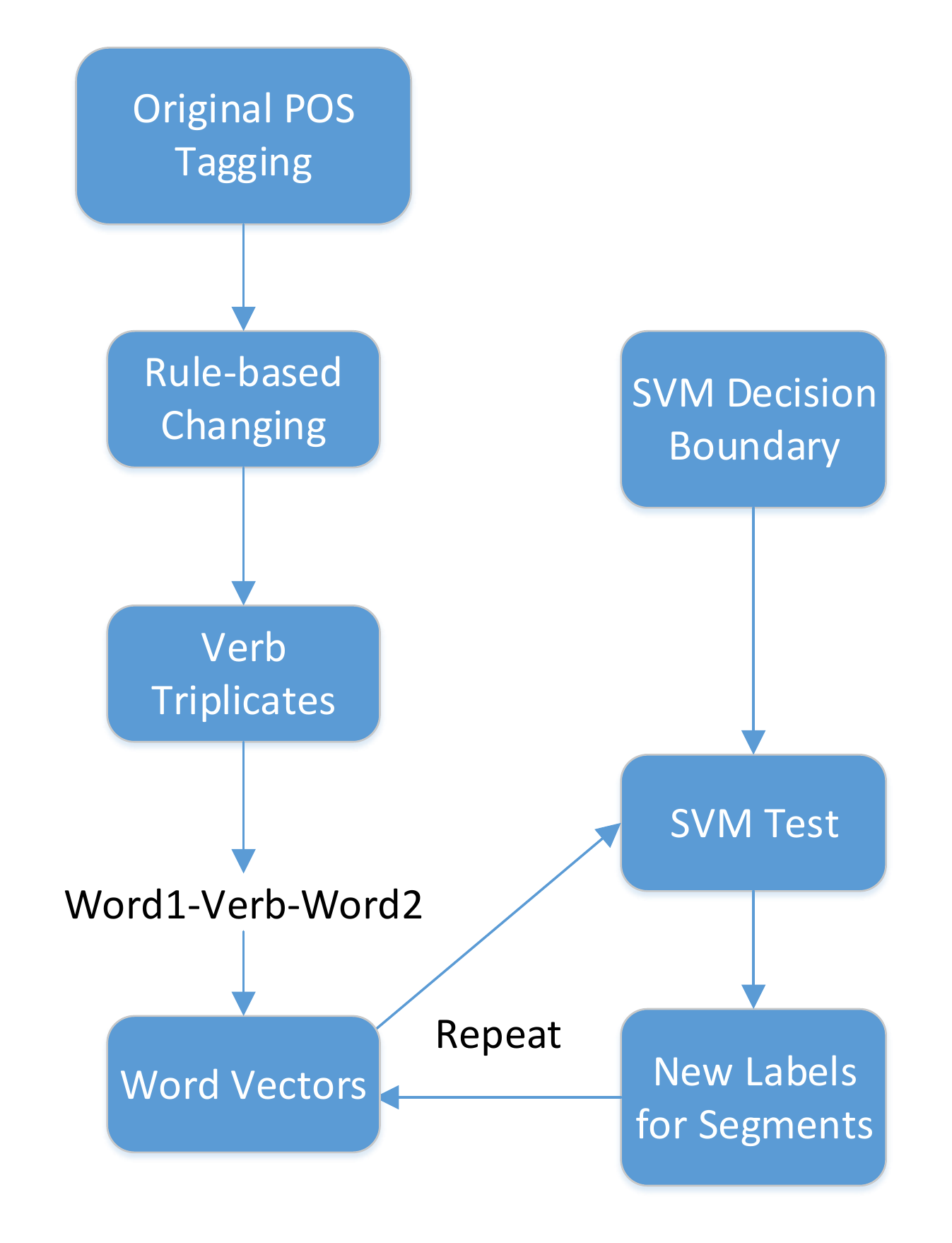}
        \caption{Flowchart of Test Stage of Automatic POS Tag Corrector}
        \label{fig:TestingStage}
    \end{subfigure}
    \caption{Training and Testing Flowchart of Automatic POS Tag Corrector}
    \label{fig:ACModel}
\end{figure*}

With the corpus of correct POS tags collected from our AMT campaign in hand, our focus shifted to designing a system to provide natural language processing features from patent claims depicted in Fig. \ref{fig:sysDiag}.  This system combines the ordinary Stanford NLP parser with a POS tag corrector trained via machine learning.  After an initial run of the ordinary Stanford NLP parser, the tag corrector determines whether or not the tags of words labelled as verbs are correct, and if not, changes them to either noun, adjective, or adverb tags.  In our prototype system, this tag corrector is the concatenation of a simple rule based corrector, which corrects the most commonly occurring tagging errors in patent claims listed in Table \ref{tab:Rules}, followed by a support vector machine trained off of the curated dataset from the previous section.  Next, Stanford NLP is run on the segment a second time, but this time the corrected POS tags are entered as extra constraints.  The process could potentially be repeated multiple times if necessary, but our initial experiments have only utilized one stage of correction.

Bearing this in mind, the remainder of this section describes the training and operation of the tag correction system.  Sections \ref{sec:tagErr}, \ref{sec:treeErr}, and \ref{sec:subjClass} will describe the evaluation of the resulting system via comparison of the tags it produces with the tags created by turkers, the use of standard NLP metrics comparing the resulting parse trees, and the improvement in patent subject classification enabled by the new patent claim NLP features, respectively.

\subsection{Training an Automatic Tag Corrector}\label{sec:train}

Fig. \ref{fig:TrainingStage} shows the flowchart of the training algorithm for the automatic tag corrector.  First of all, a rule-based corrector is applied to automatically change the tag of any word in the corpus that we had previously determined to be a particular tag, the list of which is included in Table \ref{tab:Rules}.  Following that, verb triplicates, which are trigrams centered around the verbs-in-question, are sorted into whether the putative verbs-in-question, i.e. those terms that were originally labelled as verbs by the unmodified Stanford parser, were tagged as nouns, adjectives, adverbs, or actually verbs by the turkers.  

\begin{table}[ht!]
\footnotesize
\centering
\begin{tabular}{| l | c | c |}
\hline
Word & Parts of Speech & Tag \\
\hline
Said & Adjective & JJ \\
Means & Noun & NN \\
Claim & Noun & NN \\
Predetermined & Adjective & JJ \\
Wherein & Adverb & WRB \\
According & Adverb & RB \\
\hline
\end{tabular}
\caption{Rules Used in Rule Based System}
\label{tab:Rules}
\end{table}

As the collected corpus from the previous section contains only on the order of hundreds of thousands of terms, many terms that will need to be tagged by the system are unlikely to appear in the training corpus.  However, a method of training the tag corrector to figure out what to do for these words is necessary.  In order to map words with similar meanings into comparable feature vectors, we decided to represent the verb triplicates in this step by using concatenated word vectors generated by word2vec \cite{word2vec}.  Word vectors map words to a vector space in such a way that similar words are spaced closer together than dissimilar words.  By creating these word vectors, we are able to train a POS tag corrector based on features that were in a space where related phrases are closer together and thus may be classified under the same POS class.  The feature vectors are the concatenation of the word vector for each word in the verb triplicate, created using word2vec \cite{word2vec}.  Finally a basic multiclass linear-kernel SVM training algorithm was run using SVMTorch \cite{SVMTorch}, with a class for each of nouns, adjectives, adverbs, and verbs.  In the rest of the paper, we will indicate which parts of the AMT curated dataset was used to train this multiclass SVM, as it varies between the subject classification experiments, where we leverage the whole AMT dataset to train the tag corrector, and the NLP evaluations where must hold out some of the AMT dataset for testing.

\begin{figure*}[t!]
    \centering
    \begin{subfigure}[t]{1\textwidth}
        \centering
        \includegraphics[width=0.9\textwidth]{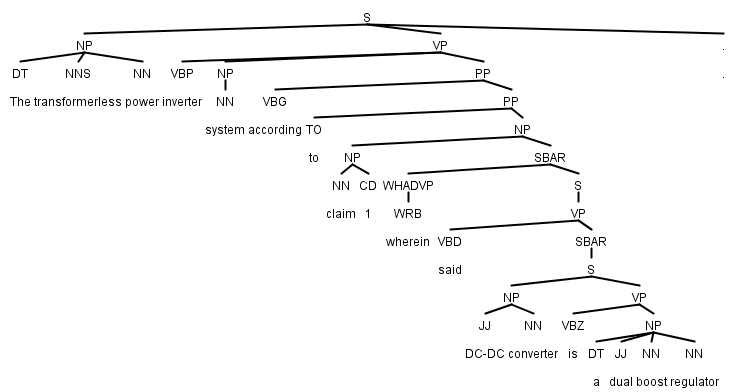}
        \caption{Original Parse Tree}
        \label{fig:OriginalParse}
    \end{subfigure}%
    
    \begin{subfigure}[t]{1\textwidth}
        \centering
        \includegraphics[width=0.9\textwidth]{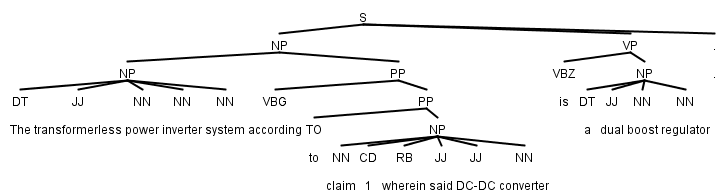}
        \caption{Parse Tree with Only SVM-based Verb Corrector}
        \label{fig:SVMParse}
    \end{subfigure}
    
    \begin{subfigure}[t]{1\textwidth}
        \centering
        \includegraphics[width=0.9\textwidth]{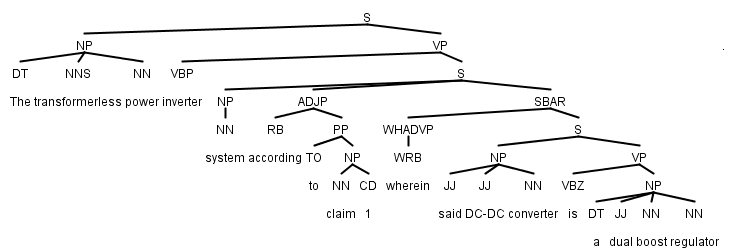}
        \caption{Parse Tree with Only Rule-based Verb Corrector}
        \label{fig:RuleParse}
    \end{subfigure}
    
    \begin{subfigure}[t]{1\textwidth}
        \centering
        \includegraphics[width=0.9\textwidth]{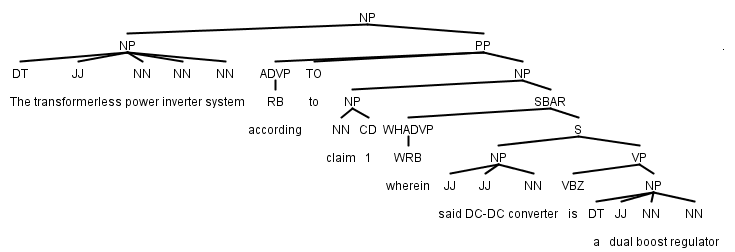}
        \caption{Parse Tree with Both Verb Correctors}
        \label{fig:BothParse}
    \end{subfigure}
    \caption{Different Parse Trees Created Using Different Methods of Correcting Verbs}
    \end{figure*}

An example series of parse trees produced through the stages of this improved claim NLP system is depicted in Fig.s \ref{fig:OriginalParse}-\ref{fig:BothParse}.  The remainder of the paper is dedicated to quantifying the improvement in patent claim parsing enabled by this system. 

\section{Experimental Validation: Improved Claim NLP}
\label{sec:eval}

In this section, we aim to validate that the claim natural language processing system described and trained in the previous section leads to improved parsing of patent claims over existing unmodified natural language processing software, and measure the improvement garnered.  Two methods of validation are provided.  The first, and simplest, approach simply compares the output of the automatic tag corrector with the tags of the ``verbs'' in question on the remainder of the curated AMT tag dataset not yet utilized to train the tag corrector, then measures the percentage of tags that the Automatic Tag Corrector correctly fixes.  The second approach forms parse trees from the original Stanford Parser, sentences with tags forced by the results from AMT, and the output sentences of the Automatic Tag Corrector and measures the difference between them using standard NLP metrics.

\subsection{Comparing POS Tags}\label{sec:tagErr}
We tested our system using our data curated according to the work in Section \ref{Curation}.  For this experiment, we only used the data from the last iteration of our AMT campaign since we noticed that the responses were more accurate.  One third of the data was used for training and two thirds were used for testing.  Altogether 113,995 words originally tagged as verbs were used in the experiment with 39,097 used for training and 74,898 were used for testing.

\begin{table}[ht!]
\footnotesize
\centering
\begin{tabular}{| l | c |}
\hline
System & Percent Error \\
\hline
Stanford POS Tagger & 31.18\% \\
Rule-Based Corrector & 23.81\% \\
SVM-Based Corrector & 9.17\% \\
\hline
\end{tabular}
\caption{Results of Two Correctors versus Original Tagging}
\label{tab:CorrectorExperiments}
\end{table}

Assuming the AMT data was ground truth, our system only had a 9.17\% error in determining the true tag as can be seen in Table \ref{tab:CorrectorExperiments}.  Two systems were used to compare to the automatic corrector - the original Stanford POS Tagger and the pure rule based corrector used in the first step of our system (where the rules are given in Table \ref{tab:Rules}).  In both cases, the SVM-Based Corrector vastly outperformed either system in the case of retagging words.

\subsection{Comparing Parse Trees}\label{sec:treeErr}
Of further interest is the improvement of the overall natural language processing system, e.g. for generating parse trees or dependency lists, after the POS tags have been corrected.  Of difficulty here is the fact that no ground truth is available, as a corpus of ground truth human annotated parse trees or dependency lists for patent claims is unavailable.  One can, however, measure the difference between the original tree (i.e. before the tags from the automatic corrector are forced) and the fixed tree resulting from the improved claim processing system.

\clearpage

Additionally, a surrogate for ground truth trees can be created by taking the portion of the AMT corpus that was not utilized in training the tag corrector, and forcing its tags in the parse tree generation software.  Then both the original unmodified Stanford NLP tree (i.e. without the tags from the AMT corpus forced) and the tree generated by forcing the tags from the  automatic corrector can be compared to this surrogate ground truth to provide a method for measuring improvement.

When comparing different parse trees for the same segment, we use a standard performance measurement used in NLP systems: the precision and recall for constituents.  Constituents, as defined in Eq. \ref{eq:constituents} are groups of words that may be as a single unit or phrase \cite{Jurafsky}.  Each node on the parse tree is associated with a category of a grammar.  
Each node on the parse tree is also a subtree root, and the subtree will span a group of words (leaves) in the whole parse tree.  The starting position and end position of the spanned words are referred as start and finish respectively \cite{Manning}. Thus we can list all the constituents of a parse tree of certain sentence.  Given a sentence, two parse trees can be compared by counting the number of common constituents with precision and recall being defined in the following equation.

\begin{align}
\label{eq:constituents}
constituent = [Node\;label, start, finish] \\
precision = \frac{|P_{constituent}\cap T_{constituent}|}{|P_{constituent}|} \\
recall = \frac{|P_{constituent}\cap T_{constituent}|}{|T_{constituent}|}
\end{align}

\begin{figure}[t]
\centering
\includegraphics[width=0.8\textwidth]{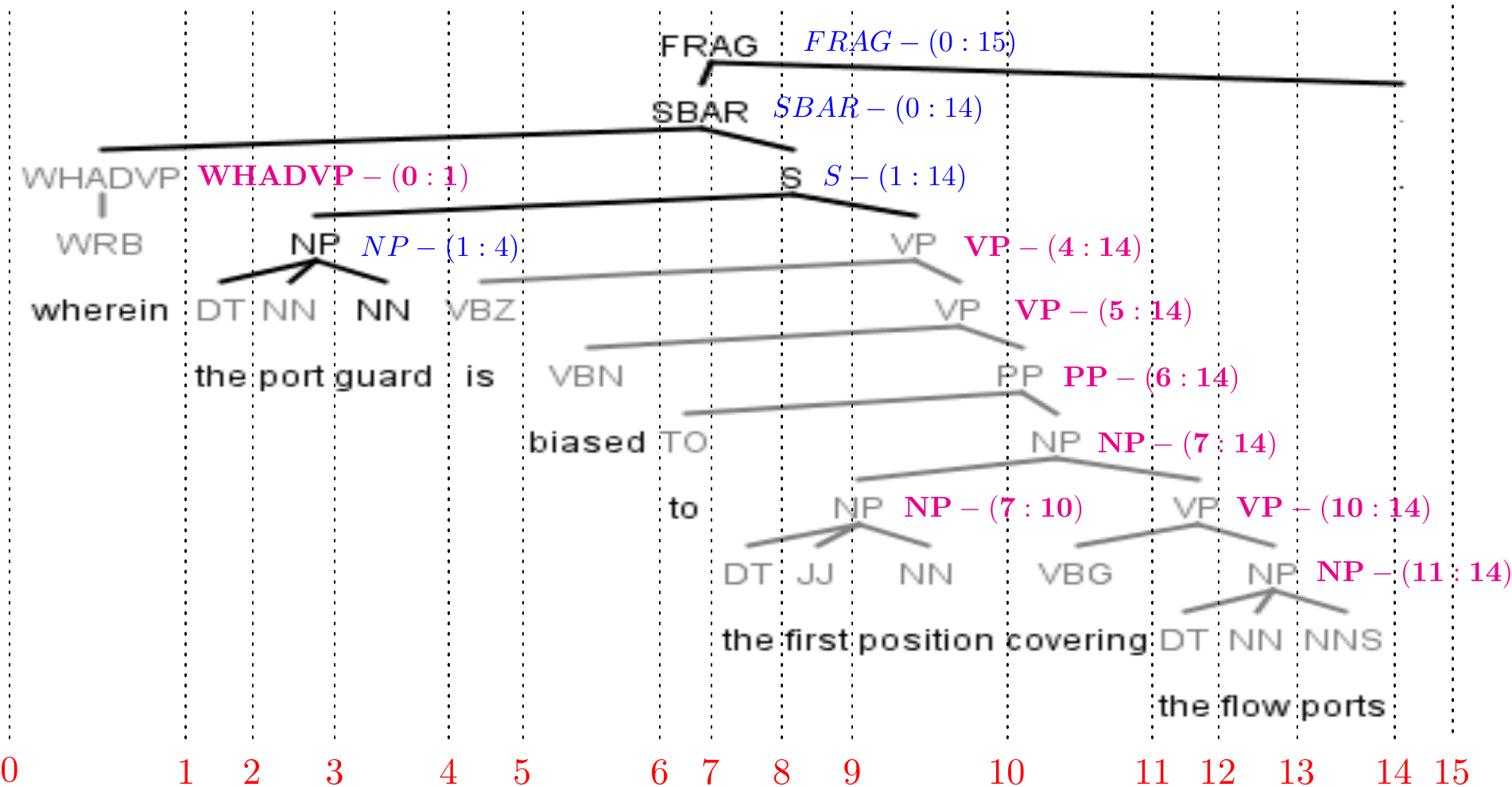}
\caption{Golden Tree: correct parse tree}
\label{fig:correctTree}
\end{figure}

\begin{figure}[t]
\centering
\includegraphics[width=0.8\textwidth]{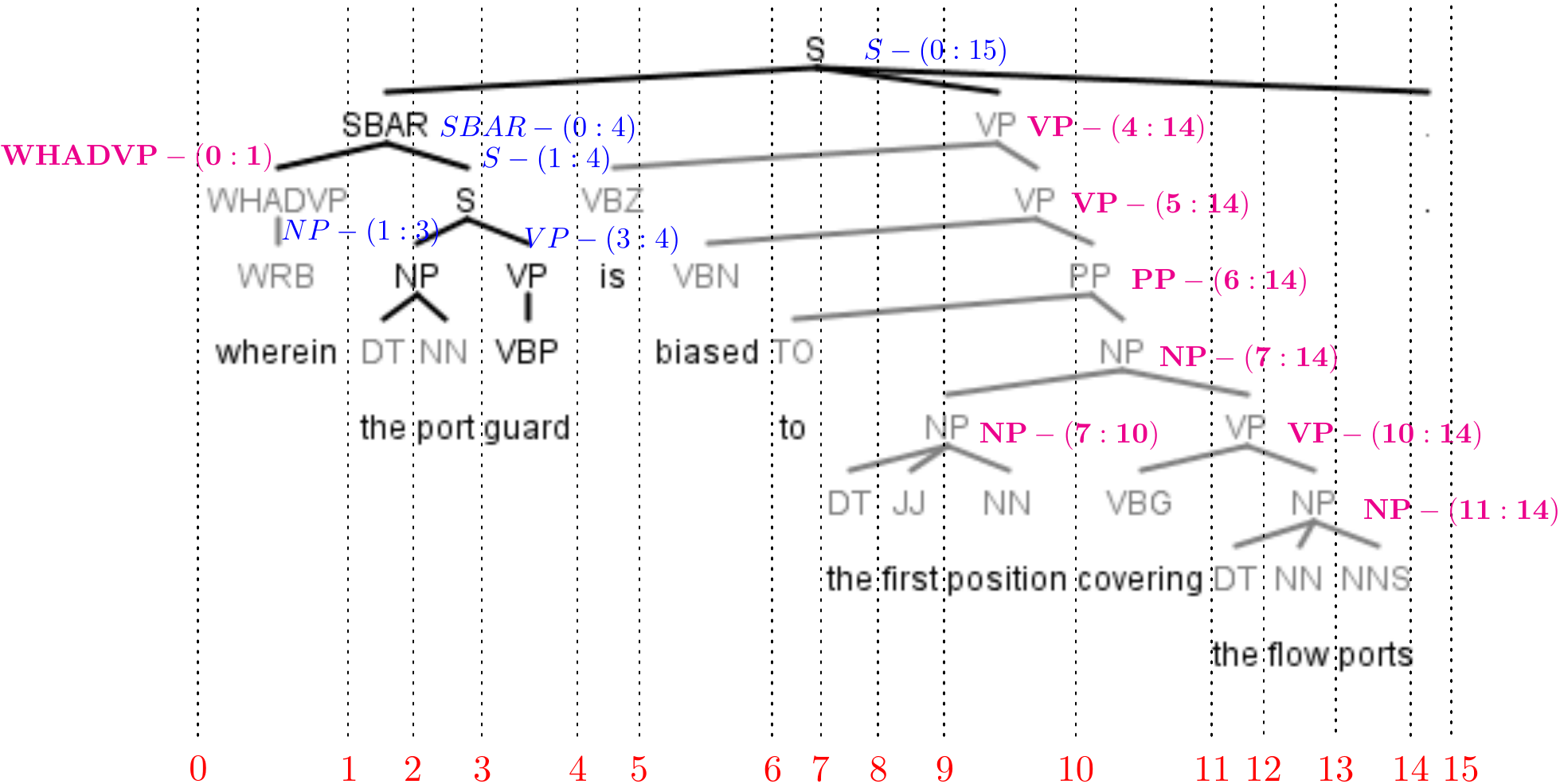}
\caption{Parse Tree from Parser}
\label{fig:wrongTree}
\end{figure}

This precision and recall measurement process is depicted with an example in Fig.s \ref{fig:correctTree} and \ref{fig:wrongTree}, for which the precision and recall are $\frac{8}{12}$ and $\frac{8}{13}$ respectively. 

\begin{table*}[t!]
\footnotesize
\centering
  \begin{tabular}{| c | c | c || c |}
    \hline
    & \multicolumn{2}{| c ||}{Patent Treebank from AMT as golden $G$} & \multicolumn{1}{c |}{Tree from Automatic Corrector as golden $G$ } \\ \cline{2-4}
    & Stanford Parser $P_{1}$ & Automatic Corrector Parser $P_{2}$ & Stanford Parser \\
    \hline
    LP &66.55\%  &82.90\% &58.11\%\\ 
    \hline 
    LR &71.80\%  &84.49\% &60.59\%\\ 
    \hline
    F-1 &69.08\% &83.69\% &59.32\% \\ 
    \hline
  \end{tabular}
  \caption{Comparing Parse Trees using Tasks 1,2,3 for training and Task 4 for testing}
  \label{tab:123,4}
\end{table*}

\begin{table*}[t!]
\footnotesize
\centering
  \begin{tabular}{| c | c | c || c |}
    \hline
    & \multicolumn{2}{| c ||}{Patent Treebank from AMT as golden $G$} & \multicolumn{1}{c |}{Tree from Automatic Corrector as golden $G$ } \\ \cline{2-4}
    & Stanford Parser $P_{1}$ & Automatic Corrector Parser $P_{2}$ & Stanford Parser \\
    \hline
    LP &70.33\%  &81.39\% &70.58\%\\ 
    \hline 
    LR &73.68\%  &77.32\% &75.01\%\\ 
    \hline
    F-1 &71.91\% &79.31\% &72.72\% \\ 
    \hline
    \end{tabular}
  \caption{Comparing Parse Trees using Task 4 for both Train and Test}
  \label{tab:4,4}
\end{table*}

\begin{table*}[t!]
\footnotesize
\centering
  \begin{tabular}{| c | c | c || c |}
    \hline
    & \multicolumn{2}{| c ||}{Patent Treebank from AMT as golden $G$} & \multicolumn{1}{c |}{Tree from Automatic Corrector as golden $G$ } \\ \cline{2-4}
    & Stanford Parser $P_{1}$ & Automatic Corrector Parser $P_{2}$ & Stanford Parser \\
    \hline
    LP &66.89\%  &87.24\% &64.93\%\\ 
    \hline 
    LR &71.74\%  &89.13\% &68.16\%\\ 
    \hline
    F-1 &69.23\% &88.17\% &66.50\% \\ 
    \hline
  \end{tabular}
  \caption{Comparing Parse Trees using Tasks 1,2,3 for both Train and Test}
  \label{tab:123,123}
\end{table*}

Three different experiments were performed which compared the parse trees from the Stanford Parser ($P_{1}$) and the automatic corrector parse trees ($P_2$) with the AMT-annotated parse trees.  These three differed depending on the portion of the data used for train and the portion used for testing.  For one experiment (recorded in Table \ref{tab:123,4}), the model used to correct segments was trained using the first three Turk Tasks and then it was tested on the fourth Turk Task.  For the second experiment (recorded in Table \ref{tab:4,4}), both training and testing came from segments in the fourth Turk Task, but the segments were separated based off of their originating patent.  For the third experiment (recorded in Table \ref{tab:123,123}), both training and testing came from the first three experiments mixed together, with training and testing being split like with the second experiment.

As can be seen from Tables \ref{tab:123,4}, \ref{tab:4,4}, and \ref{tab:123,123}, the results show that the Automatic Corrector's parse tree is much closer to the AMT's parse tree than the unmodified original Stanford NLP parse tree.  Both precision and recall are at least 4\% higher when comparing with the Automatic Corrector than with the regular Stanford Parser.  In addition, as could be seen, comparing the Stanford parse with either the AMT parse or the Automatic Corrector parse produces similar results too, suggesting that there might be a fundamental difference in the way that the Stanford Parser performs without any side-information on the parts of speech tags of the patent claims.

\section{Experimental Validation: Improved Patent Subject Classification}
\label{sec:subjClass}

The experiments in the previous section have provided evidence supporting the hypothesis that the augmented patent claim NLP system trained from the AMT garnered dataset is more effective at parsing sentences from patent claims than the original unmodified grammar distributed with the Stanford NLP system it builds upon.  Because ground truth parse trees were not available, however, all that could be measured was differences between the original parse trees and those obtained with parts of speech created through the automatic correcting system and the AMT labels.  The large differences demonstrated do indicate improvement, however, a better validation of the improved claim NLP system would show that the features it provides are useful in other patent processing systems.

To show that our system has merit in this sense, a toy problem in the field of patent subject classification was created to compare the performance of dependencies generated with just the Stanford Parser to those created after being run through our automatic POS tag fixer.  This problem, subject classification, was chosen in particular both because it is well posed and because reliable ground-truth labels from which to train and test a classifier are easily gathered, as these subject classification labels are available freely as part of the published patent documentation.

\subsection{Background on Patent Subject Classification}

The patent subject classification utilized by the United States Patent and Trademark Office has evolved substantially over the years.  Patents submitted to the United States Patent and Trademark Office historically were categorized using at least one of three different classification systems: the United States Patent Classification (USPC) system, the International Patent Classification (IPC) system, and the Cooporative Patent Classification (CPC) system.  The USPC system was developed and maintained by the USPTO in order to classify patents entering in the patent office as well as providing search restrictions when performing a prior art search \cite{USPTO_MPEP}.  In an attempt to unify the classification of international patents among different patent offices as well as to provide a singular classification system for prior art search, the IPC system was developed by World Intellectual Property Organization (WIPO).  This system was only meant to supplement and not replace any particular classification system \cite{IPCGuide}.  More recently, the CPC system, based off of the European Classification system which itself was a more specific and detailed version of the IPC, was developed jointly by the European Patent office (EPO) and USPTO in order to provide a singular world patent classification used by many offices as the primary classification system \cite{CPCWebsite}.

While the layouts between the USPC and IPC/CPC are different, the two systems are based off of a hierarchical classification scheme.  The USPC has two main levels, class level and subclass level, with over 150,000 subclasses.  Each subclass may also be a child of other subclasses in the same class (e.g. Subclass 714/748 is a child of Subclass 714/746 which itself is a child of Subclass 714/699).  In the IPC/CPC system, each patent is given a classification featuring at least 5 different hierarchical levels: Section, Class, Subclass, Group, and Main Group.  Patents may also be given a subgroup level classification.  Subgroups are children of main groups or, similar to the USPC, other subgroups.  Altogether, there are over 72,000 groups and subgroups that patents could be classified in under the IPC and 250,000 different groups and subgroups under the CPC.

Early work on automated subject classification focussed on the USPC,  \cite{larkey1997some,[276]larkey1999patent}.  However, after the development of the WIPO-alpha patent classification database, which  collected patents gathered via the Patent Cooperation Treaty which were classified using the IPC system \cite{[8]CJFall_AutoCategoryInternationalPatentClassif_2003_SIGIR}, focus shifted to the more distinctly hierarchical IPC and CPC.  Due to the overwhelming number of subgroups in the IPC, most researchers tend to focus on classifying patents on the subclass level \cite{[270]beney2010lci,[271]derieux12010combining,[272]guyot2010myclass,[273]verberne2011patent}.  This limits the number of classes to under 700.  In addition, due to the hierarchical nature of the classification system, some researchers may focus on developing hierarchical classifiers to classify on the subclass or possibly main group level \cite{[267]seeger2006cross,[268]tikk2003experiment,[274]cai2004hierarchical,[278]rousu2006kernel} or create tasks to classify at the subgroup level while knowing details about higher levels \cite{CLEFIP2011}.  Very few papers attempt to classify at the subgroup level of the IPC without knowing details about higher levels \cite{[263]chen2012three}.


\subsection{Experimental Setup}

While a full fledged patent subject classification experiment at scale is well beyond this scope of this article, the aim of this section is rather to show with a representative, but tiny, example that the new features the article provides can improve patent subject classification.  In particular, it has already been demonstrated \cite{[273]verberne2011patent} that dependency features can provide helpful information when performing subject classification.  Our goal here is to show that the improved patent claim NLP system provides dependencies which can yield substantial improvement in subject classification.

\begin{figure}[h]
\centering
\includegraphics[width=.9\textwidth]{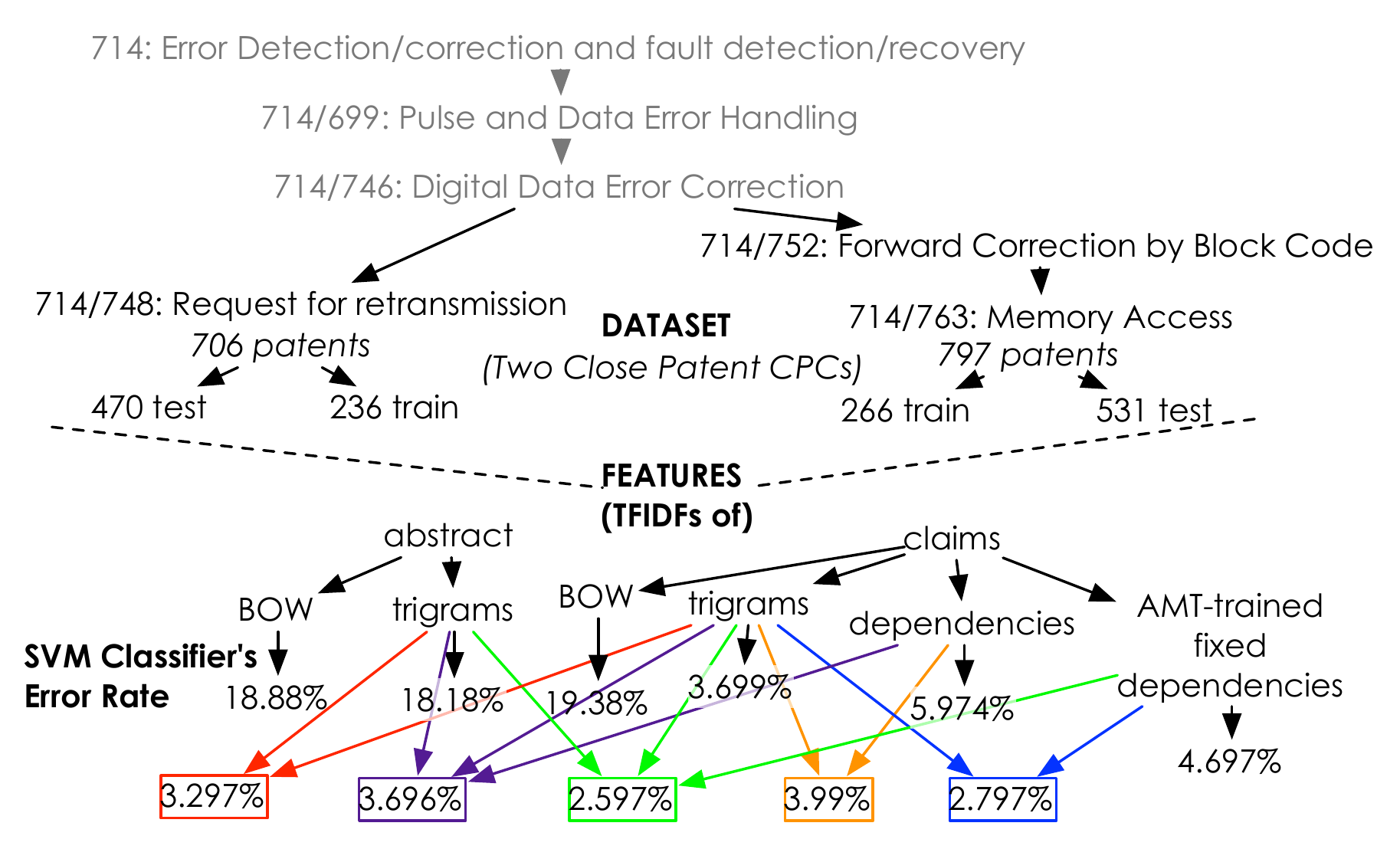}
\caption{A simple small patent subject classification problem between two closely related subject classes.  Consistently the improved claim NLP features provided by the system trained and described in this paper decrease  automated subject classification error rates, while the same features provided by the original unmodified NLP system worsen them.}\label{fig:classificationExample}
\end{figure}

To do this, a very simple patent subclass classification experiment was set up.  In order to keep the experiment small, but difficult, we selected two subclasses of the same parent class  \emph{714: Error detection/correction and fault detection/recovery} of the USPC.  As depicted in Fig. \ref{fig:classificationExample}, the two subclasses selected, \emph{714/748: Request for retransmission} and \emph{714/763: Memory Access}, are both refinements of the subclass \emph{714/746: Digital Data Error Correction}, which is in turn a refinement of \emph{714/699: Pulse and Data Error Handling}.  These subclasses were selected because they are clearly closely related (owing to their common parent subclasses), and in fact, because there are patents which take both classification labels.  The precise subject classification problem selected was to correctly identify the primary classification label of a series of presented ``test'' patent based on a series of different groups of features.  Note that the labels are only given off the primary subclass, and that there are patents that are also classified under the other class as a secondary classification.


All patents published with one of these two classifications (714/748 and 714/763) as their primary classification between January 1, 1976 and August 21, 2015 were collected, and each patent was assigned to either a testing set or training set.  The number of patents used for training and testing are depicted in Fig. \ref{fig:classificationExample}.

As the focus of the experiment is demonstrating the power of the new features, multiple systems were built by using the same baseline type of classifier, a SVM \cite{SVMTorch}, on different types of features.  The features were used were all variants of term frequency inverse document frequency feature vectors (TF-IDF) generated using Scikit-Learn \cite{scikit-learn}, but with the notion of ``terms'' varying between: 1) individual words in the document -- called a bag of words (BOW) model, 2) three consecutive words in the document (trigrams), and 3) an entire Stanford dependency triplicate -- two words appearing in a sentence in the document together with one of several possible relationships between them \cite{StanfordDependencies}.  Furthermore, these TF-IDFs were aggregated over either the abstract of a patent, or a patent's claims, or both.  In the case of patent claims, both the dependencies generated by unmodified language model provided with Stanford NLP, and those generated with the improved claim NLP system described in this paper were utilized.  Finally, systems were also created by concatenating these different types of feature vectors.

\subsection{Experimental Results}

The classification error rates provided by the classifiers trained from the various feature types are depicted in Fig. \ref{fig:classificationExample}.  A first observation this figure enables, in line with common sense, is that the information provided in patent claims is consistently more important than that provided in the abstract.  More importantly, however, this figure demonstrates that in multiple contexts, the improved claim NLP system provided in this paper provides additional information that improves the classification error rate.  For instance, using the claim dependencies alone in subject classification between these two classes, the improved claim NLP system decreases the error rate from 5.974\% to 4.695\%.  A system built from claim trigrams alone can achieve an error rate of 3.699\%, and when these features are augmented with those garnered by the language model provided with the unmodified Stanford NLP, the error rate actually worsens to 3.99\%, while augmenting the claim trigrams with claim dependencies built from the improved claim NLP system in this paper instead decreases the the error rate to 2.797\%.  Similarly, a system that incorporates both features built from abstract trigrams and claim trigrams can achieve an error rate of 3.297\%.  If one augments these feature vectors with the claim dependencies provided by unmodified Stanford NLP, one worsens the error rate again to 3.696\%, while if one augments the same features with the claim dependencies provided by the improved claim NLP system described in this paper, one obtains an error rate of 2.597\%.  Clearly, the increase in accuracy of the dependencies provided by the improved claim NLP system is consistently providing more useful information for subject classification.

\section{Conclusion}
As they describe the legal scope an invention, patent claims are arguably the most important section of a patent for many patent processing problems.  Natural language processing features built from these claims could potentially be useful for software to aid humans in multiple stages of the patent processing pipeline.  However, claims are not written in the same style of English as appears in news articles and on the web, and, as such, the language model provided with off the shelf natural language processing software is not well suited to claims.  Gathering a treebank of annotations for patent claims from which to train a completely new language model for patent claims would be an expensive and exhausting exercise, requiring hard to find expertise.  Motivated, however, by the observation that much of the difficulty in parsing claims lies in the mis-identification of verbs, this paper proposed an alternative way of adapting NLP software to claim language by simply correcting the POS tags of putative verbs.  In particular, we described the collection, via Amazon Mechanical Turk, and subsequent curation of a dataset providing part of speech tags for terms in patent claims mistaken by standard language models as verbs.  This dataset was leveraged to train a POS tag corrector, which, when combined with the Stanford Natural Language Processing system, provided an improved system for patent claim NLP with far less skilled effort and expense than would have been required to train a completely new model for claim language.  The utility of the improved claim NLP system was demonstrated by showing both the substantial changes and improvement in the parse trees it provides over the original unmodified system, and the decrease in subject classification error rate enabled by using the improved features.

\bibliographystyle{Fullname}
\bibliography{PatentMining}
\end{document}